\documentclass[twoside,leqno,twocolumn]{article}
\usepackage{ltexpprt}
\usepackage{amssymb}
\usepackage{graphicx}
\usepackage{amsmath}
\usepackage{algorithm}
\usepackage{algpseudocode}
\usepackage{subfigure}
\usepackage{booktabs}

\begin{document}

\title{\Large Generative Adversarial Classifer for Handwriting Characters Super-Resolution}
\author{Zhuang QIAN\thanks{Department of Electrical and Electronic Engineering of Xi'an Jiaotong-Liverpool University}\\
\and
Kaizhu HUANG\footnotemark[1]\\
\and
Qiufeng WANG\footnotemark[1] \\
\and
Jimin XIAO\footnotemark[1] \\
\and
Rui ZHANG\thanks{Department of Mathematical Sciences of Xi'an Jiaotong-Liverpool University}
}
\date{}

\maketitle








\begin{abstract} \small\baselineskip=9pt Generative Adversarial Networks (GAN) receive great attentions recently due to its excellent performance in image generation, transformation, and super-resolution. However, GAN has rarely been  studied and trained for classification, leading that the generated images may not be appropriate for classification. In this paper, we propose a novel Generative Adversarial Classifier (GAC) particularly for low-resolution Handwriting Character Recognition. Specifically, involving additionally a classifier in the training process of   normal GANs, GAC is calibrated for learning suitable structures and restored characters images that benefits the classification. Experimental results show that our proposed method can achieve remarkable performance in handwriting characters $8\times$ super-resolution, approximately 10\% and 20\% higher than the present state-of-the-art methods respectively on benchmark data  CASIA-HWDB1.1 and MNIST.
	
\end{abstract}

\section{Introduction.}The super-resolution (SR) which estimates a high-resolution (HR) image from its low-resolution (LR) counterpart is a highly important task in computer vision. SR has attracted much attention from the field of computer vision research and has a wide range of applications~\cite{chowdhuri2012very,nasrollahi2014super,yang2007spatial}.

Convolutional neural networks have achieved excellent performance in super-resolution, however there are two main challenges. One is that the high-frequency information lacked in LR image cannot be reconstructed very well. The early neural network can get a good HR image from given an LR image at small scale factors by minimizing  the mean squared error (MSE) between the reconstructed image and the ground truth~\cite{srcnn,fsrcnn,VDSR,DRCN,ESPCN}. However, these methods may fail to reconstruct high quality images at large scale factors such as 4$\times$. The deep networks such as DRRN~\cite{DRRN}, EDSR~\cite{EDSR} and MDSR~\cite{EDSR} can achieve high PSNR in reconstructed image however the details of the high-frequency information are missing. The other challenge is that reconstructed details of images are fabricated. SRGAN~\cite{srgan} is good at restoring high-frequency information of HR images while the PSNR is relatively low because some of high-frequency information is fabricated and is unfaithful to the ground-truth.

\begin{figure}[h]
	\centering
	\subfigure[Bicubic]{
		\includegraphics[width=0.4\textwidth]{{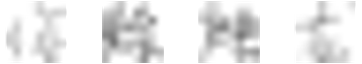}}
		\label{subfig:bicubic_result}}	
    \subfigure[SRResNet~\cite{srgan}]{\includegraphics[width=0.4\textwidth]{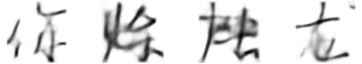}
		\label{subfig:resnet_result}}
	\subfigure[SRGAN~\cite{srgan}]{
	\includegraphics[width=0.4\textwidth]{{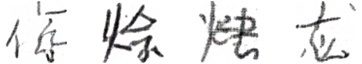}}
		\label{subfig:SRGAN_result}
	}
	\subfigure[The proposed GAC]{ \includegraphics[width=0.4\textwidth]{{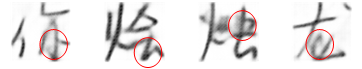}}
		\label{subfig:cgan_result}
	}
	\subfigure[Ground-truth]{
		\includegraphics[width=0.4\textwidth]{{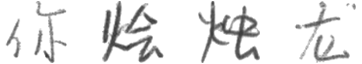}}
		\label{subfig:HR}
	}
	\caption{The reconstructed image of $8\times$ scale} \label{fig:result}
\end{figure}

For characters, the details are important, since details often determine whether characters can be recognisable. For example, the characters in the 
Figure~\ref{subfig:bicubic_result}-Figure~\ref{subfig:SRGAN_result} are difficult to recognise, and some are mis-recognised. The lack of high-frequency information in general deep networks and the fake high-frequency details of the GAN make details become obstacles to
recognise for both computer and human. In another word, these networks would not be appropriate for characters super-resolution.
Therefore, it is necessary to propose a new network which is suitable for characters super-resolution.

In this paper, we propose a novel network based on SRGAN~\cite{srgan}, removing the VGG loss while adding a classifier module to classify images reconstructed by the generator. There are some reconstructed images given by various methods in Figure~\ref{fig:result}. We highlight some pixels in the red cycle to show the difference with the results from SRGAN~\cite{srgan} and SRResNet. The proposed method called Generative Adversarial Classifier (GAC) can reconstruct images with more high-frequency information than SRResNet and more faithful to groun-truth than SRGAN~\cite{srgan}.

Using the classification loss as additional information so that we can constrain the generator and make the reconstructed images more recognisable. In this sense, our network is similar to Triple-GAN~\cite{triplegan}. Triple-GAN also has three parts where the discriminator will decide whether a pair of image and its label $\left ( x,y \right )$ come from the true distribution $p\left ( x,y \right )$. This distribution discrimination model makes Triple-GAN unable to deal with data with large number of classes (e.g., CASIA-HWDB 1.1 containing 3,755 classes~\cite{dataset}), since input image and 3,755 classes label into discriminator will occupy a large amount of memory. In contrast, in our network, the discriminator and classifier are only associated with generator, the discriminator does not distinguish the distribution of images and labels, which makes our network: (1) suitable for the problem with large number of classes in particular for Chinese character recognition, (2) much easier to optimise because the parameter number in our discriminator is far fewer than Triple-GAN's even if our discriminator is significantly deeper.

The $8 \times$ scale reconstructed image recognition rate of our network is $10\%$ higher than SRGAN on CASIA-HWDB1.1~\cite{dataset} with a 8 upscaling factor. In comparison, the top-1 accuracy is 63.95\% and top-3 accuracy is 80.69\%, the top-1 accuracy and top-3 accuracy of SRGAN is 53.28\% and 69.52\% respectively. Besides the CASIA-HWDB 1.1, we also evaluated our proposed methods on benchmark data MNIST~\cite{mnist} and CIFAR-10~\cite{cifar} \footnote{it is not a handwriting dataset}, and the experimental results show our method can achieve significantly better results than the present state-of-the-art approaches.

\begin{figure}
	\centering
	\includegraphics[height=5cm]{{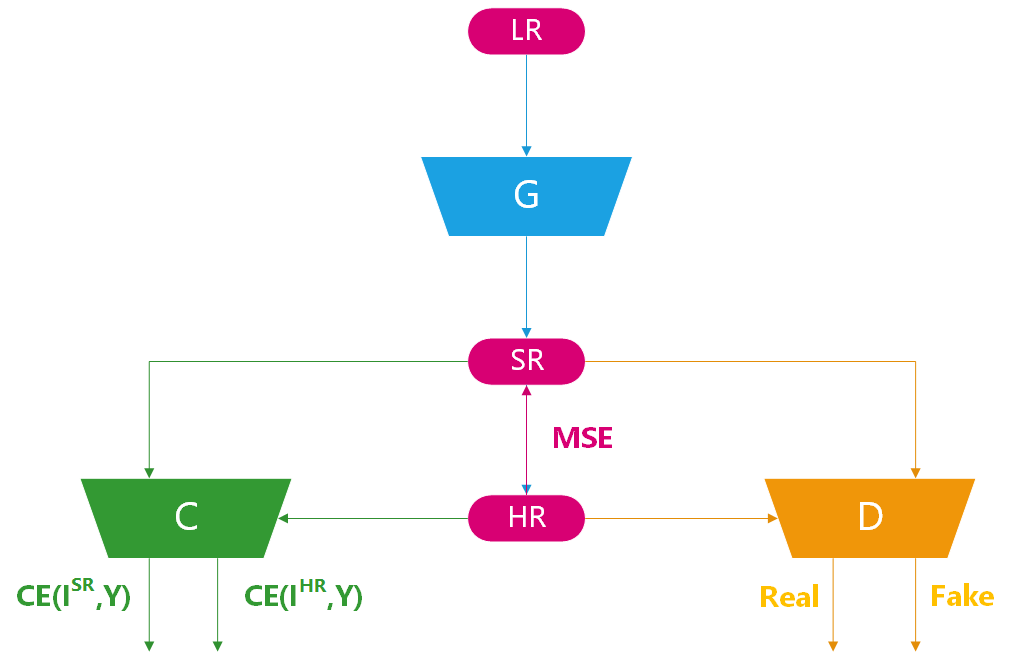}}
	\caption{
		An illustration of our GAC (best viewed in color). The utilities of D, C and G are colored
		in yellow, green and blue respectively, with
		“CE” denoting the cross entropy loss for supervised learning. The generator generate the reconstructed image $I^{SR}$, classifier label the true image $I^{HR}$ and $I^{SR}$ by minimizing cross entropy loss of $\left(I^{HR},Y\right)$ and $\left(I^{SR},Y\right)$, the discriminator distinguishes the whether images come from the true images.}
	\label{fig:framework}
\end{figure}

\section{Related Work}
In this section, we will conduct an overall review of related work, including image super-resolution and Generative Adversarial Nets.
\subsection{Image Super-resolution}
The research of image super-resolution can be divided into two categories: one is based on single image super-resolution (SISR), and the other is based on multiple image super-resolution (MISR)~\cite{areview}. Our work can be cast into the first category.
We will focus on single image super-resolution (SISR) and will not further discuss approaches that reconstruct HR images from multiple images.

Recently, convolutional neural network (CNN) based SR algorithms have shown excellent performance. In Wang et al.~\cite{wang2015deep}, the authors encoded a sparse representation prior into a feed-forward network architecture based on the learned iterative shrinkage and thresholding algorithm (LISTA)~\cite{gregor2010learning}. Dong et al.~\cite{srcnn2,srcnn} used bicubic interpolation to downscale an image as input image and trained a three layer convolutional network end-to-end. The deeply-recursive convolutional network (DRCN)~\cite{DRCN} is a highly effective architecture that allows long-range pixel dependencies while keeping the number of model parameters small. Johnson et al.~\cite{johnson2016perceptual} and Bruna et al.~\cite{bruna2015super} proposed a perceptual loss function to reconstruct visually more convincing HR images.

\subsection{Generative Adversarial Nets (GAN)}

Generative Adversarial Nets (GAN) is proposed by Goodfellow~\cite{gan} which contains two parts, a generator and a discriminator. The generator is responsible for generating images close to the real pictures to fool the discriminator, and the discriminator is responsible to discriminate the picture from the generator or real pictures. Adversarial examples problem is also proposed and there are many methods to solve it such as~\cite{adversericalexam}.

 In 2016, Radford et al.~\cite{dcgan} proposed DCGAN which is stable in most settings and shows the vector arithmetics as an intrinsic property of the representations learned by the Generator. Mirza et al.~\cite{conditional} proposed the conditional GAN, the idea is to use labels for some data to help network build salient representations, it can control the generator's outputs without changing the architecture by adding label as another input to the generator. Ledig et al.~\cite{srgan} proposed the SRGAN by reconstructing the HR image with GAN based on Resnet~\cite{resnet} and it achieves remarkable performance in human vision but low PSNR. The Triple-GAN is proposed by Li et al.~\cite{triplegan} which contains three parts, a classifier $C$ that (approximately) characterizes the conditional distribution $p_{c}\left ( y\mid x \right )\approx p\left ( y\mid x \right )$, a class-conditional generator $G$ that (approximately) characterizes the conditional distribution in the other direction $p_{g}\left ( x\mid y \right )\approx p\left ( x\mid y \right )$, and a discriminator
$D$ that distinguishes whether a pair of data $\left ( x; y \right )$ comes from the true distribution $p\left ( x; y \right )$, the final goal of Triple-GAN is to predict the labels $y$
for unlabeled data as well as to generate new samples $x$ conditioned on $y$.

\section{Method}

\begin{figure}
	\centering
	\subfigure[Generator architecture (G in Figure~\ref{fig:result})]{
		\includegraphics[width=0.4\textwidth]{{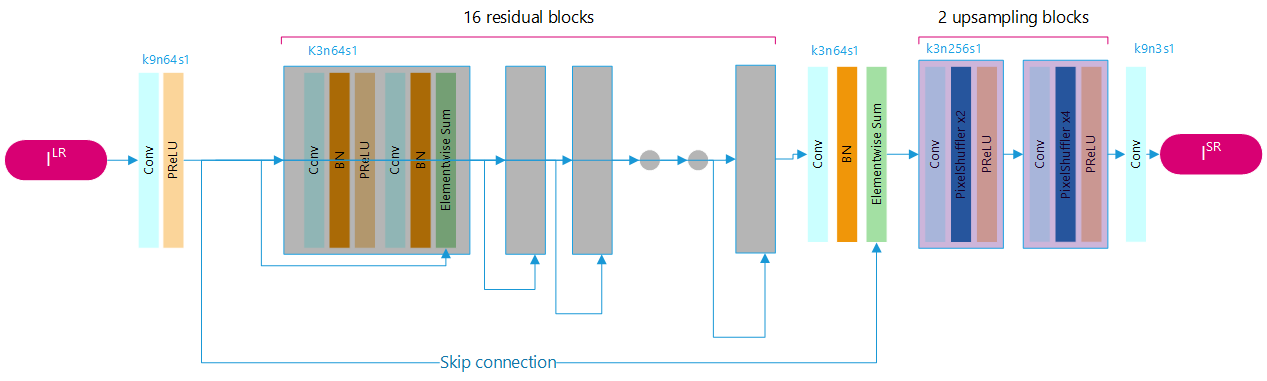}}
		\label{subfig:G}
	}
	\subfigure[Discriminator architecture (D in Figure~\ref{fig:result})]{
		\includegraphics[width=0.4\textwidth]{{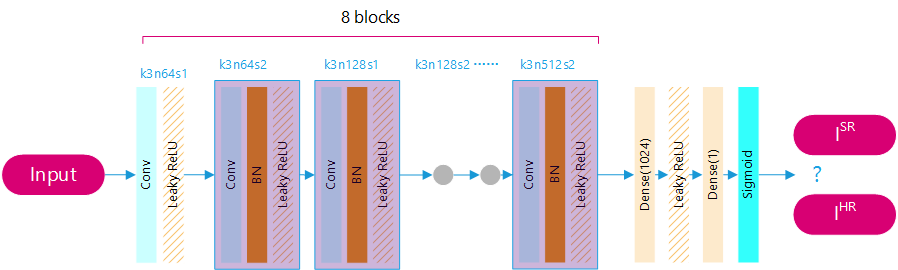}}
		\label{subfig:D}
	}
	\subfigure[Classifier architecture (C in Figure~\ref{fig:result})]{
		\includegraphics[width=0.4\textwidth]{{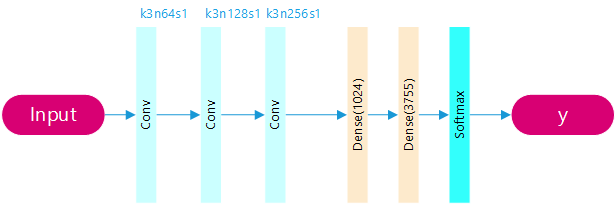}}
		\label{subfig:C}
	}
	
	\caption{GAC architecture} \label{fig:architecture}
	
\end{figure}

In single image super-resolution, the aim is to estimate a high-resolution $I^{SR}$ from a low-resolution input image $I^{LR}$. Here the $I^{LR}$ is the low-resolution image of  its high-resolution counterpart $I^{HR}$. In our network, there are labels for  $I^{HR}$.  The proposed overall network can be illustrated in Figure~\ref{fig:framework}. The generator $G$ generates the reconstructed images $I^{SR}$ from given low-resolution images $I^{LR}$, the discriminator $D$ distinguishes the $I^{SR}$ from $I^{HR}$, and the classifier $C$ gives labels for  $I^{SR}$. The discriminator $D$ and classifier $C$ are both linked to the generator $G$, trying to guide  the generator $G$ for generating more realistic yet recognisable reconstructed $I^{SR}$ images.

Our ultimate goal is to train a generating function $G$ that
estimates a reconstructed $I^{SR}$ image as good as possible for a given LR input image. To achieve this, we train a generator network as a CNN $G_{\theta _{G}}$ parametrized $\theta _{G}$. For training images $I_{n}^{HR}$, $n= 1,....,N$ with corresponding $I_{n}^{LR}$, $n= 1,....,N$, the SR-specific problem is formulated as:

\begin{equation}
\widehat{\theta }_{G}= arg \min \limits_{\theta_{G}}\frac{1}{N}\sum_{n=1}^{N}l^{SR}\left ( G_{\theta _{G}}\left ( I_{n}^{LR} \right ),I_{n}^{HR} \right )
\end{equation}

In this work we will specifically design a loss function
$l^{SR}$ as a weighted combination of several loss components.

\subsection{Adversarial Network Architecture}

Inspired by Goodfellow et al.~\cite{gan} and SRGAN~\cite{srgan}, we define a discriminator network $D_{{\theta}_{D}}$ which we optimize alternately with the generator $G_{{\theta}_{G}}$, and the optimized object is to solve the adversarial
min-max problem:
\begin{equation}
\begin{aligned}
\min\limits_{G_{\theta}}\max\limits_{D_{\theta}}     ~&\mathbb{E}_{I^{HR}\sim p_{GT}\left(I^{HR}\right)}\left[\log D_{\theta_{D}}\left(I^{HR}\right)\right]+\\
&\mathbb{E}_{I^{LR}\sim p_{G}\left(I^{LR}\right)}\left[\log \left(1-D_{\theta_{D}}\left(G_{\theta_{G}}\left(I^{LR}\right)\right)\right)\right]
\end{aligned}
\label{gan-eq}
\end{equation}

This formulation follows the basic working principle of GAN. It trains a generator model $G$ to try to fool a
discriminator $D$ which is trained to distinguish
super-resolved images from real images. With this approach
the generator can learn to reconstruct image more realistic and  highly similar to real images, even can make discriminator difficult to discriminate true images from reconstructed images.
This approach encourages the result of generator perceptually superior in human vision, and it can achieve preferable visual perception, compared to the traditional method obtained by minimizing pixel-level error measurements such as the Mean Square Error(MSE).

For our generator network $G$ and discriminator network $D$, we exploit the SRGAN architecture~\cite{srgan}. The generator network illustrated in Fig.~\ref{subfig:G} are 16 residual blocks with identical layout where the block consists two convolutional layers with small $3\times3$ kernels and 64 feature maps followed by batch-normalization layers~\cite{batchnor} and Parametric ReLU~\cite{PReLU} as the activation function. Before the output layer, We increase the resolution of the image with two trained upsampling blocks which contain one convolutional layer with small $3\times3$ kernels followed by one sub-pixel convolution layer~\cite{subpixel} with $scale=2~or~4$ and Parametric ReLU as the activation function.

For the discriminator network $D$ illustrated in Fig.~\ref{subfig:D}, we follow the architectural guidelines summarized by Radford et al.~\cite{dcgan}. We engage the LeakyReLU
activation $(\alpha= 0:2)$ and avoid max-pooling throughout
the network. The discriminator network is trained to solve
the maximization problem in Equation~\ref{gan-eq}. It contains eight
convolutional layers with an increasing number of $3\times3$
filter kernels, increasing by a factor of 2 from 64 to 512 kernels. Strided convolutions are used to reduce the image resolution each time the number of features is doubled. The resulted 512 feature maps are followed by two dense layers and a final sigmoid activation function.

For the classifier, we simply apply it with 3 convolutional layers with an increasing number of $3\times3$ filter kernels, increasing by a factor of 2 from 64 to 128 kernels followed by 2
two dense layers and a final softmax activation function to obtain a probability for sample classification as illustrated in Figure~\ref{subfig:C}.
\subsection{Loss Function}

The definition of loss function $l^{SR}$ is critical
for the performance of our generator network. While
$l^{SR}$ is commonly modeled based on the MSE~\cite{srcnn}, we
design a loss function that assesses a solution with respect
to perceptually relevant characteristics. We formulate the
loss function as the weighted sum of a content loss and an adversarial loss component and a classification loss component as:
\begin{equation}
l^{SR}=l^{SR}_{mse}+10^{-3}\cdot l^{SR}_{adv}+\alpha\cdot l^{SR}_{cla}
\end{equation}

\subsubsection{Content Foss}

We use the pixel-wise MSE loss as our content loss calculated as:
\begin{equation}
l^{SR}_{mse}=
\dfrac {1}{rWH}\sum ^{rW}_{x=1}\sum ^{rH}_{y=1}\left( I^{HR}_{x,y}-G_{\theta _{G}}\left( I^{LR }\right) _{x,y}\right) ^{2}
\end{equation}
where $l^{LR}$ and $l^{HR}$ is the low-resolution image and high-resolution image respectively, $W$, $H$, $C$ and $r$ is the width, height, channel and scale factor, respectively. We describe $l^{LR}$ by a real-valued tensor of size $W\times H \times C$ and $l^{HR}$, $l^{SR}$ by $rW\times rH \times C$ respectively. For character images, $C$ can be set to 1 or 3 generally.

This is the most widely used optimization target for image
super-resolution. However, while achieving particularly high PSNR, solutions of MSE optimization problems often lack high frequency content; this results in perceptually unsatisfying
solutions with overly smooth textures.

\subsubsection{Adversarial Loss}
Following the GAN architecture, we add the adversarial loss to our loss function. This encourages our network to generate images more natural and realistic in vision, by trying to fool the discriminator network. The adversarial loss $l_{adv}^{SR}$
is defined based on the probabilities of the discriminator over all training samples as:
\begin{equation}
l_{adv}^{SR}=\sum^{N}_{n=1}-\log D_{\theta_{D}}\left( G_{\theta_{G}} \left( I^{SR}_n\right)\right)
\end{equation}
Here, $D_{\theta_{D}}\left( G_{\theta_{G}} \left( I^{SR}_n\right)\right)$ is the probability that the reconstructed
image $G_{\theta_{G}} \left( I^{SR}_n\right)$ is judged as a natural image by the discriminator. For better
gradient behavior we minimize this equation instead
of the original GAN adversarial loss $\log\left[1-D_{\theta_{D}}\left( G_{\theta_{G}} \left( I^{SR}_n\right)\right)\right]$

\subsubsection{Classification Loss}
We introduce the third player, i.e., the classifier, into our proposed GAC model, which can characterize the conditional distribution $p_c(y|x) \approx  p(y|x)$ in general network. In our network, the classifier can label correctly for a given reconstructed image, which can be denoted as $C_{\theta_{C}} \left( I^{SR}_n\right) \approx  y$. We can achieve this simply by minimizing the cross entropy loss as:
\begin{equation}
l^{SR}_{cla}=\sum^{N}_{n=1}\left(-y_n\log\left( C_{\theta_{C}} \left( I^{SR}_n\right)\right)\right)
\end{equation}

In order to make sure that the distribution $(I^{SR},Y)$ be as close as possible to the true data distribution $(I^{HR},Y')$, we need another one loss $\boldsymbol{R}_c$ as:
\begin{equation}
\boldsymbol{R}_c=\sum^{N}_{n=1}\left(-y_n\log\left( C_{\theta_{C}} \left( I^{HR}_n\right)\right)\right)
\end{equation}

Consequently, we define the overall loss function as:
\begin{equation}
l^{SR}=l^{SR}_{mse}+10^{-3}\cdot l^{SR}_{adv}+\alpha\cdot l^{SR}_{cla}+\boldsymbol{R}_c
\label{eq:GAC}
\end{equation}

The detailed algorithm to minimize the overall loss function is given in Algorithm~\ref{alg:Framwork}.

\begin{algorithm}[!h]
	\caption{Minibatch stochastic gradient descent training method of Classified-GAN in SSL}
	\label{alg:Framwork}
	\begin{algorithmic}
		\For{number of training iterations}
		\begin{itemize}
			\item[$\bullet$] Sample a batch of pairs $(I^{LR};I^{HR}; Y)$  of size $N$
			
			\item[$\bullet$] Update $D$ by ascending along its stochastic gradient:
			\begin{displaymath}
			\widehat{\theta }_{D}\left[ \frac{1}{N} \left( \sum _{n}^N\log D\left( I^{HR}_n\right)- \log D\left( G_{\theta_{G}} \left( I^{SR}_n\right)\right)  \right)  \right]
			\end{displaymath}
			
			\item[$\bullet$]Update $C$ by descending along $\boldsymbol{R}_c$ stochastic gradient:
			\begin{displaymath}
			\widehat{\theta }_{C}\left[ \frac{1}{N} \left( \sum _{n}^N- y_n\log\left( C_{\theta_{C}} \left( I^{HR}_n\right)\right)\right)\right]
			\end{displaymath}
			
			\item[$\bullet$]Update $C$ by descending along its stochastic gradient:
			\begin{displaymath}
			\widehat{\theta }_{C}\left[ \frac{1}{N} \left( \sum _{n}^N-y_n\log\left( C_{\theta_{C}} \left( I^{SR}_n\right)\right) \right)  \right]
			\end{displaymath}
			
			\item[$\bullet$]Update $G$ by descending along its stochastic gradient:
			\begin{displaymath}
			\begin{aligned}	
			\widehat{\theta }_{G} \left[ \frac{1}{N}
			\left( \sum _{n}^N \left( I^{HR}_{n}-G_{\theta _{G}}\left( I_n^{LR }\right) \right) ^{2} \right. \right. \\
			\left. \left. - \log D\left( G_{\theta_{G}} \left( I^{SR}_n\right)\right) -y_n\log\left( C_{\theta_{C}} \left( I^{SR}_n\right)\right)\right)  \right]	
			\end{aligned}		
			\end{displaymath}

		\end{itemize}
		\EndFor\State
	\end{algorithmic}
\end{algorithm}

\section{Experiments}
\subsection{Experimental Set-up}
We perform experiments on the widely used handwriting Chinese
characters dataset CASIA-HWDB1.1~\cite{dataset}, handwriting digits dataset MNIST~\cite{mnist}. In addition, to further check if our method could work well in non-text data, we also evaluate our methods on CIFAR-10~\cite{cifar}.

CASIA-HWDB1.1 consists of 897,758 training samples, and 223,991 testing samples for 3,755 classes. On CASIA-HWDB1.1 and CIFAR-10 datasets, experiments are performed with a scale factor of $8\times$ between low- and high-resolution
images from $8\times8$ to $64\times64$. On MNIST, we set that scale factor is $7\times$ from $4\times4$ to $28\times28$. We also implemented other super-resolution methods  include bicubic, SRResNet\cite{resnet}, SRGAN~\cite{srgan} and Triple-GAN~\cite{triplegan} and compared them on the three benchmark datasets. Our code will be uploaded to GitHub once this paper is published.

We obtained the LR images
by downsampling the HR images using bicubic kernel with downsampling scale factor $r = 8$ on CASIA-HWDB1.1 and CIFAR-10, and $r = 7$ on MNIST. For each mini-batch we pick 128 random  HR images
of distinct training images. Note that we can apply the
generator model to images of arbitrary size as it is fully
convolutional. The MSE loss
was thus calculated on images of intensity range $[-1; 1]$. We employed the trained MSE-based SRResNet
network as initialization for the generator when training
the actual GAN to avoid undesired local optima.

As we mentioned above, the performance of the network is not easily measured by the human eyes. For character recognition, the simplest performance test method is  to recognise the reconstructed character image with to the classifier and exploit the recognition accuracy as the evaluation standard. We train a simple classifier $C_{0}$ for measurement, which contains 3 Convolutional Layers and 2 dense layers.  $C_{0}$ can achieve the 89.29\% in top 1 accuracy on CASIA-HWDB1.1.

For the SRResNet and SRGAN, we first train these two networks on CASIA-HWDB1.1 training set by using the $64\times64$ images as HR images and downsampling $8\times$ these images to  $8\times8$ as input. Then we get the test reconstructed images by downsampling  CASIA-HWDB1.1 test set to $8\times8$ as input. Finally, we can use the $C_{0}$ to recognise the reconstructed test images.

For our proposed GAC model, we use two strategies to train it. The first strategy is to initialize $C$ to $C_{0}$, then freeze $C$ network so that its parameters are not updated. In this way, $C$ plays the same role in the network as VGG used in SRGAN, restricting the distribution $(I^{SR},Y)$ trending to the ground-truth distribution $(I^{HR},Y')$.  The second strategy is to initialize $C$ to $C_{0}$, and $C$ network updated its parameters during training. In this strategy, it makes the distribution $(I^{SR},Y)$ deviate from the ground-truth distribution $(I^{HR},Y')$, but $C$ becomes more suitable for generator. The hyper-parameter $\alpha$ in Equation~\ref{eq:GAC} was tuned empirically and we choose the best one on a validation set.

\subsection{Experimental Results}

 We report the experimental results on CASIA-HWDB1.1. in Table~\ref{Table:HWDB}. As clearly observed,  our proposed GAC model achieves significantly better performance than all the comparison algorithms.  In particular, the GAC model without fixing $C$ achieves the top-1 accuracy of $63.95\%$, around $10.7\%$ higher than SRGAN, the best of the other competitive algorithms. On the other hand, a simplified version of GAC with fixed $C$ also leads to significant improvement over SRGAN. Note that, on CASIA-HWDB1.1, we did not report the performance of Triple-GAN since it is  intractable to be trained on the large category data CASIA-HWDB1.1 with $3,755$ classes due to its inherit nature.
 
\begin{table}
	\centering
	\caption{Recognition accuracy of reconstructed test images on CASIA-HWDB1.1}
	\label{Table:HWDB}
	\begin{tabular}{ccc}	
		\toprule  
		Method & top-1(\%)& top-3(\%)\\
		\midrule  
		bicubic & 2.24 & 4.04\\
		SRResNet & 36.33 & 50.73\\
		SRGAN & 53.28 & 69.52\\
		Triple-GAN & - & - \\
		GAC(fixed $C$, $\alpha$=0.001) & 58.24 & 74.03\\
		\textbf{GAC($\alpha$=0.0005)} & \textbf{63.95} & \textbf{80.69}\\			
		HR & 89.29 & 95.86\\
		\bottomrule 
	\end{tabular}
\end{table}

\begin{figure}
	\centering
	\includegraphics[height=4cm]{{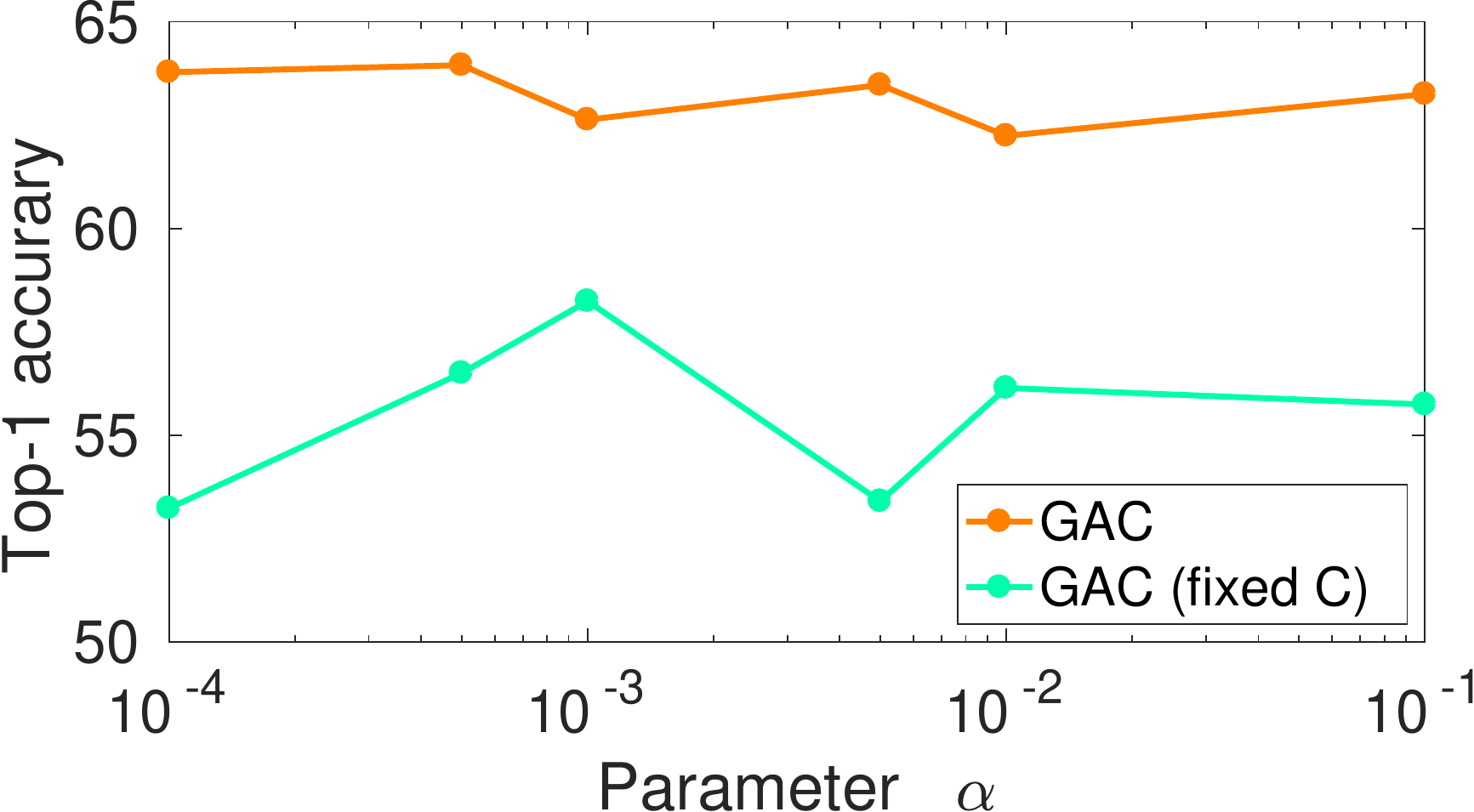}}
	\caption{The top-1 accurary of different $\alpha$ in CASIA-HWDB1.1}
	\label{fig:parameter}
\end{figure}

To further check the sensitivity of the proposed GAC on the hyper-parameter $\alpha$ as defined in Equation~\ref{eq:GAC}, we also report the recognition performance against different $\alpha$ on CASIA-HWDB1.1. These results can be seen in Figure~\ref{fig:parameter}. We can observe that, though the proposed GAC network is insensitive to the hyper-parameter $\alpha$ in general, smaller values may usually lead to better performance. In contrast, GAC (fixed C) is more sensitive to $\alpha$ than GAC, and the smaller $\alpha$ is, the more network close to SRGAN. 

On MNIST and CIFAR-10 we also apply both two above-mentioned strategies, but do not need to initialize $C$ to $C_{0}$, since $\boldsymbol{R}_c$ is enough to train the model well for 10-class datasets. For Triple-GAN, since its original purpose is not suitable for super-resolution, we remove the label from input of the generator so as to adapt to our task. The results are reported in Table~\ref{Table:first}. From Table~\ref{Table:HWDB} and~\ref{Table:first}, we can see that three components combined training including our proposed GAC model and Triple-GAN can improve the recognition accuracy substantially. Furthermore, our proposed GAC model outperforms Triple-GAN with a  $19.62\%$ and $15.33\%$ higher accuracy respectively on MNIST and CIFAR-10. These results validates the effectiveness of the proposed GAC model.

\begin{table}
	\centering
	\caption{Recognition accuracy of reconstructed test images on MNIST and CIFAR-10}
	\label{Table:first}
	\begin{tabular}{ccc}	
		\toprule  
		Dataset & MNIST(\%)& CIFAR-10(\%)\\
		\midrule  
		bicubic & 12.17 & 10.00\\
		SRResNet & 36.75 & 10.66\\
		SRGAN & 42.99 & 11.06\\
		Triple-GAN & 74.07 & 37.28 \\
		GAC(fixed $C$, $\alpha$=0.001) & 93.50 & 42.68\\
		\textbf{GAC($\alpha$=0.001)} & \textbf{93.69} & \textbf{53.61}\\		
		HR & 98.91 & 62.14\\
		\bottomrule 
	\end{tabular}
\end{table}

\section{Conclusion}
We propose a new three-player generative adversarial classifier (GAC) with three components, a generator, a discriminator and a classifier, particularly for the purpose of character super-resolution. Specifically, involving additionally a classifier in the training process of   normal GANs, GAC is calibrated for learning suitable structures and restored characters images that benefits the classification.  Our empirical results on CASIA-HWDB1.1, MNIST, CIFAR-10 datasets
demonstrate that GAC can achieve the state-of-the-art
classification results for character super-resolution.

\section*{Acknowledgement}
The work was partially supported by the following: National Natural Science Foundation of China under no.~61473236 and 61876155; The Natural Science Foundation of the Jiangsu Higher Education Institutions of China under no. 17KJD520010; Suzhou Science and Technology Program under no. SYG201712, SZS201613; Natural Science Foundation of Jiangsu Province BK20181189, 17KJB520041; Key Program Special Fund in XJTLU under no. KSF-A-01, KSF-A-10, KSF-P-02.

\bibliography{bibliography}

\begin{thebibliography}{10}

\bibitem{areview}
{\sc S.~Borman and R.~L. Stevenson}, {\em Super-resolution from image
  sequences-a review}, in Circuits and Systems, 1998. Proceedings. 1998 Midwest
  Symposium on, IEEE, 1998, pp.~374--378.

\bibitem{bruna2015super}
{\sc J.~Bruna, P.~Sprechmann, and Y.~LeCun}, {\em Super-resolution with deep
  convolutional sufficient statistics}, arXiv preprint arXiv:1511.05666,
  (2015).

\bibitem{triplegan}
{\sc L.~Chongxuan, T.~Xu, J.~Zhu, and B.~Zhang}, {\em Triple generative
  adversarial nets}, in Advances in Neural Information Processing Systems,
  2017, pp.~4091--4101.

\bibitem{chowdhuri2012very}
{\sc D.~Chowdhuri, K.~Sendhil~Kumar, M.~R. Babu, and C.~P. Reddy}, {\em Very
  low resolution face recognition in parallel environment}, IJCSIT)
  International Journal of Computer Science and Information Technologies, 3
  (2012), pp.~4408--4410.

\bibitem{srcnn2}
{\sc C.~Dong, C.~C. Loy, K.~He, and X.~Tang}, {\em Learning a deep
  convolutional network for image super-resolution}, in European Conference on
  Computer Vision, Springer, 2014, pp.~184--199.

\bibitem{srcnn}
\leavevmode\vrule height 2pt depth -1.6pt width 23pt, {\em Image
  super-resolution using deep convolutional networks}, IEEE transactions on
  pattern analysis and machine intelligence, 38 (2016), pp.~295--307.

\bibitem{fsrcnn}
{\sc C.~Dong, C.~C. Loy, and X.~Tang}, {\em Accelerating the super-resolution
  convolutional neural network}, in European Conference on Computer Vision,
  Springer, 2016, pp.~391--407.

\bibitem{gan}
{\sc I.~Goodfellow, J.~Pouget-Abadie, M.~Mirza, B.~Xu, D.~Warde-Farley,
  S.~Ozair, A.~Courville, and Y.~Bengio}, {\em Generative adversarial nets}, in
  Advances in neural information processing systems, 2014, pp.~2672--2680.

\bibitem{gregor2010learning}
{\sc K.~Gregor and Y.~LeCun}, {\em Learning fast approximations of sparse
  coding}, in Proceedings of the 27th International Conference on International
  Conference on Machine Learning, Omnipress, 2010, pp.~399--406.

\bibitem{PReLU}
{\sc K.~He, X.~Zhang, S.~Ren, and J.~Sun}, {\em Delving deep into rectifiers:
  Surpassing human-level performance on imagenet classification}, in
  Proceedings of the IEEE international conference on computer vision, 2015,
  pp.~1026--1034.

\bibitem{resnet}
\leavevmode\vrule height 2pt depth -1.6pt width 23pt, {\em Deep residual
  learning for image recognition}, in Proceedings of the IEEE conference on
  computer vision and pattern recognition, 2016, pp.~770--778.

\bibitem{batchnor}
{\sc S.~Ioffe and C.~Szegedy}, {\em Batch normalization: Accelerating deep
  network training by reducing internal covariate shift}, arXiv preprint
  arXiv:1502.03167,  (2015).

\bibitem{johnson2016perceptual}
{\sc J.~Johnson, A.~Alahi, and L.~Fei-Fei}, {\em Perceptual losses for
  real-time style transfer and super-resolution}, in European Conference on
  Computer Vision, Springer, 2016, pp.~694--711.

\bibitem{VDSR}
{\sc J.~Kim, J.~Kwon~Lee, and K.~Mu~Lee}, {\em Accurate image super-resolution
  using very deep convolutional networks}, in Proceedings of the IEEE
  Conference on Computer Vision and Pattern Recognition, 2016, pp.~1646--1654.

\bibitem{DRCN}
\leavevmode\vrule height 2pt depth -1.6pt width 23pt, {\em Deeply-recursive
  convolutional network for image super-resolution}, in Proceedings of the IEEE
  conference on computer vision and pattern recognition, 2016, pp.~1637--1645.

\bibitem{cifar}
{\sc A.~Krizhevsky, V.~Nair, and G.~Hinton}, {\em The cifar-10 dataset},
  online: http://www. cs. toronto. edu/kriz/cifar. html,  (2014).

\bibitem{mnist}
{\sc Y.~LeCun, C.~Cortes, and C.~Burges}, {\em Mnist handwritten digit
  database}, AT\&T Labs [Online]. Available: http://yann. lecun.
  com/exdb/mnist, 2 (2010).

\bibitem{srgan}
{\sc C.~Ledig, L.~Theis, F.~Husz{\'a}r, J.~Caballero, A.~Cunningham, A.~Acosta,
  A.~Aitken, A.~Tejani, J.~Totz, Z.~Wang, et~al.}, {\em Photo-realistic single
  image super-resolution using a generative adversarial network}, arXiv
  preprint,  (2016).

\bibitem{EDSR}
{\sc B.~Lim, S.~Son, H.~Kim, S.~Nah, and K.~M. Lee}, {\em Enhanced deep
  residual networks for single image super-resolution}, in The IEEE Conference
  on Computer Vision and Pattern Recognition (CVPR) Workshops, vol.~1, 2017,
  p.~3.

\bibitem{dataset}
{\sc C.-L. Liu, F.~Yin, D.-H. Wang, and Q.-F. Wang}, {\em Online and offline
  handwritten chinese character recognition: benchmarking on new databases},
  Pattern Recognition, 46 (2013), pp.~155--162.

\bibitem{adversericalexam}
{\sc C.~Lyu, K.~Huang, and H.-N. Liang}, {\em A unified gradient regularization
  family for adversarial examples}, in Data Mining (ICDM), 2015 IEEE
  International Conference on, IEEE, 2015, pp.~301--309.

\bibitem{conditional}
{\sc M.~Mirza and S.~Osindero}, {\em Conditional generative adversarial nets},
  arXiv preprint arXiv:1411.1784,  (2014).

\bibitem{nasrollahi2014super}
{\sc K.~Nasrollahi and T.~B. Moeslund}, {\em Super-resolution: a comprehensive
  survey}, Machine vision and applications, 25 (2014), pp.~1423--1468.

\bibitem{dcgan}
{\sc A.~Radford, L.~Metz, and S.~Chintala}, {\em Unsupervised representation
  learning with deep convolutional generative adversarial networks}, arXiv
  preprint arXiv:1511.06434,  (2015).

\bibitem{ESPCN}
{\sc W.~Shi, J.~Caballero, F.~Husz{\'a}r, J.~Totz, A.~P. Aitken, R.~Bishop,
  D.~Rueckert, and Z.~Wang}, {\em Real-time single image and video
  super-resolution using an efficient sub-pixel convolutional neural network},
  in Proceedings of the IEEE Conference on Computer Vision and Pattern
  Recognition, 2016, pp.~1874--1883.

\bibitem{subpixel}
\leavevmode\vrule height 2pt depth -1.6pt width 23pt, {\em Real-time single
  image and video super-resolution using an efficient sub-pixel convolutional
  neural network}, in Proceedings of the IEEE Conference on Computer Vision and
  Pattern Recognition, 2016, pp.~1874--1883.

\bibitem{DRRN}
{\sc Y.~Tai, J.~Yang, and X.~Liu}, {\em Image super-resolution via deep
  recursive residual network}, in The IEEE Conference on Computer Vision and
  Pattern Recognition (CVPR), vol.~1, 2017.

\bibitem{wang2015deep}
{\sc Z.~Wang, D.~Liu, J.~Yang, W.~Han, and T.~Huang}, {\em Deep networks for
  image super-resolution with sparse prior}, in Proceedings of the IEEE
  International Conference on Computer Vision, 2015, pp.~370--378.

\bibitem{yang2007spatial}
{\sc Q.~Yang, R.~Yang, J.~Davis, and D.~Nist{\'e}r}, {\em Spatial-depth super
  resolution for range images}, in Computer Vision and Pattern Recognition,
  2007. CVPR'07. IEEE Conference on, IEEE, 2007, pp.~1--8.

\end{thebibliography}
\bibliographystyle{siam}


\end{document}